\documentclass[10pt,twocolumn,letterpaper]{article}

\usepackage[accsupp]{axessibility}
\usepackage{iccv}
\usepackage{times}
\usepackage{epsfig}
\usepackage{graphicx}
\usepackage{amsmath}
\usepackage{amssymb}
\usepackage{multirow}
\usepackage{subfigure}
\usepackage{booktabs} 
\usepackage{diagbox}
\usepackage{multicol} 
\usepackage{arydshln}
\usepackage{dashrule}

\newtheorem{theorem}{Theorem}
\newtheorem{lemma}{Lemma}

\newtheorem{definition}{Definition}
\newtheorem{assumption}{Assumption}
\usepackage[colorlinks,linkcolor=blue]{hyperref}

\newsavebox\CBox
\def\textBF#1{\sbox\CBox{#1}\resizebox{\wd\CBox}{\ht\CBox}{\textbf{#1}}}

\iccvfinalcopy 

\def\iccvPaperID{****} 

\ificcvfinal\fi

\begin{document}

\title{Lifelong Infinite Mixture Model Based on Knowledge-Driven Dirichlet Process}

\author{Fei Ye and Adrian G. Bors\\
Department of Computer Science, University of York, York YO10 5GH, UK\\
{\tt\small fy689@york.ac.uk,
adrian.bors@york.ac.uk}
}

\newlength\secmargin
\newlength\paramargin
\newlength\figmargin
\setlength{\secmargin}{-1.0mm}
\setlength{\paramargin}{-2.0mm}
\setlength{\figmargin}{-3.0mm}

\setlength{\abovedisplayskip}{6pt} 
\setlength{\belowdisplayskip}{6pt}
\setlength{\abovecaptionskip}{6pt}

\maketitle
\ificcvfinal\thispagestyle{empty}\fi

\begin{abstract}
Recent research efforts in lifelong learning propose to grow a mixture of models to adapt to an increasing number of tasks. The proposed methodology shows promising results in overcoming catastrophic forgetting. However, the theory behind these successful models is still not well understood. In this paper, we perform the theoretical analysis for lifelong learning models by deriving the risk bounds based on the discrepancy distance between the probabilistic representation of data generated by the model and that corresponding to the target dataset. Inspired by the theoretical analysis, we introduce a new lifelong learning approach, namely the Lifelong Infinite Mixture (LIMix) model, which can automatically expand its network architectures or choose an appropriate component to adapt its parameters for learning a new task, while preserving its previously learnt information. We propose to incorporate the knowledge by means of Dirichlet processes by using a gating mechanism which computes the dependence between the knowledge learnt previously and stored in each component, and a new set of data. Besides, we train a compact Student model which can accumulate cross-domain representations over time and make quick inferences. The code is available at \url{https://github.com/dtuzi123/Lifelong-infinite-mixture-model}.
\end{abstract}

\section{Introduction}

Lifelong learning (LLL) aims to learn successively a series of tasks from their corresponding probabilistic representations of specific databases. The objective of the lifelong learning model is to implement all learnt tasks at any given time. Modern deep learning approaches have been successful in a variety of applications including image translation \cite{ItoI_network}, image synthesis \cite{AC-GAN} and object detection \cite{OnlyLook}, but all these models face a major challenge in the performance, when applied on prior tasks, while learning multiple tasks, one after the other. This challenge is caused by catastrophic forgetting which happens when a model adapts its parameters in order to learn a new task \cite{LifeLong_review}.

\begin{figure}[http]
    \centering
	\includegraphics[scale=0.37]{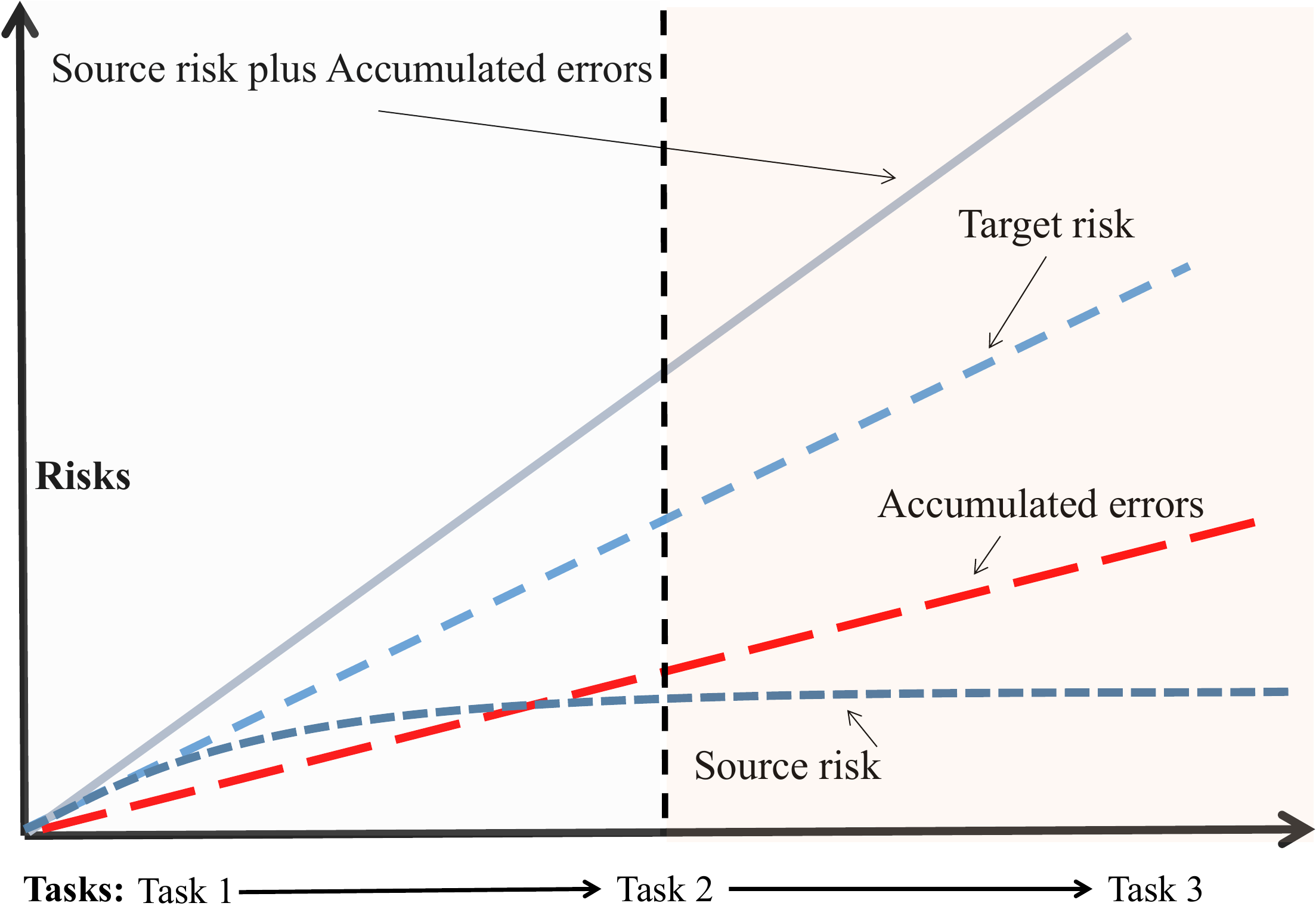}
	\caption{The process of forgetting the information from a certain dataset after learning two additional tasks. The source distribution generated by the Generative Replay Mechanims (GRM) is degenerated when learning a new task.}
	\label{theory_image}
	\vspace{-14pt}
\end{figure}

The Generative Replay Mechanism (GRM) \cite{Generative_replay} is a popular lifelong learning approach showing promising results overcoming catastrophic forgetting \cite{Lifelong_VAE,GenerativeLifelong,GenerativeConcept,MemoryGAN,LifelongTeacherStudent}. A generative replay model $G_\theta: \mathcal{Z} \to \mathcal{X}$, aims to transform a low dimensional random variable $\mathcal{Z}$ into a high dimensional variable $\mathcal{X}$. $G_\theta$ can be an implicit generative model such as a Generative Adversarial Network (GAN) \cite{GAN} or an explicit latent model such as a Variational Autoencoder (VAE) \cite{VAE}. Once a task is learnt, $G_\theta$ generates data which then can be combined with data sampled from a given database, corresponding to a new task, to form a joint dataset used for training. Some methods \cite{Lifelong_VAE,Generative_replay} reduce the memory size by only generating a batch of samples for each training step, by using only a copy $G_{\theta'}$ of the GRM. However, a major challenge for GRM based methods is that of gradually losing knowledge across tasks since a GRM model is trained on its own generations repeatedly. Another drawback, when GANs are used as GRMs, is that of facing mode collapse \cite{Veegan}. Two solutions have been proposed to address this problem. Rao {\em et al.} \cite{LifelongUnsupervisedVAE} enable GRMs with a network expansion mechanism in which the model's capacity is increased when shifting data distributions. The other solution is to use the expansion mechanism \cite{DirichletLifelong} or an ensemble structure \cite{AdversarialContinualLearning,LifelongMetaLearning,BatchEnsemble} in which each expert is built on the top of a shared module and only a single expert is updated during the training. These approaches usually preserve the best performance of previous tasks, but the theoretical analyse behind these methods is not well understood. 

In this paper, we provide the theoretical analysis for lifelong learning models, inspired by the idea illustrated in Fig.~\ref{theory_image}. The forgetting behaviour of the model, when learning a certain task, is affected by an ever increasing upper bound (solid line in Fig.~\ref{theory_image}) to the target-risk, during the lifelong leaning. This is mainly caused by the increased accumulated errors when learning additional tasks while the discrepancy between the target and source distribution is also gradually increased. However, the optimal source-risk can not guarantee a low target-risk since the source distribution of the trained model is gradually degenerated. Inspired by these results, the main idea of the proposed
Lifelong Infinite Mixture (LIMix) model is to automatically grow its network architecture if the given task is sufficiently novel when compared to the previously learned knowledge or update an appropriate component that has a small discrepancy to the given task. The Dirichlet process, which is usually computational expensive when using the expectation-maximization algorithm to estimate component parameters \cite{EMAlgorithms}, can be used for these mechanisms. In order to reduce the computational costs and make an accurate inference for the selection and expansion of the model architecture, we infer an indicator variable for each data sample by using a gating mechanism based on the Dirichlet process that computes the corrections between the knowledge stored in each component and the new data. Furthermore, by accumulating knowledge, while enabling fast inference across data domains, using a lightweight model is an attractive feature in LLL, which does not appear in existing lifelong mixtures or ensemble models \cite{DirichletLifelong,BatchEnsemble}. Our main contributions are~:

\vspace{-5pt}
\begin{itemize}
\setlength{\itemsep}{2pt}
\setlength{\parsep}{2pt}
\setlength{\parskip}{2pt}

\item We analyse the forgetting behaviour during LLL by evaluating the accumulated risk and find that the discrepancy distance between the source and target distribution is key to overcome forgetting. 

\item This is the ﬁrst study to provide theory insights when using mixture models for LLL. We also extend the theoretical analysis to explain the performance change for a model when shifting the order of tasks.

\item We propose a new lifelong mixture model with theoretical guarantees for LLL. We also explore training a compact Student model from the mixture under LLL.

\end{itemize}

\section{Related works}

Artificial lifelong learning models are trained three different approaches: regularization, dynamic architectures and memory replay. Regularization methods introduce an auxiliary term in the loss function in order to penalize changes in the network's weights when learning a new task \cite{Distilling_nets,LessForgetting,EWC,Lwf,VCL,LifeLong_combination}.  This can alleviate catastrophic forgetting but can not guarantee the effective performance on the previously learnt tasks. Dynamic architectures methods would grow the size of the network by adding processing layers or increase the number of parameters in order to adapt to a growing number of tasks \cite{Adanet,DirichletLifelong,LearnAdd,ProgressiveNN,Error_driven,OnlineLearning}. Lee \etal \cite{DirichletLifelong} used Dirichlet processes for expanding their network architecture. However, their approach does not provide any theoretical guarantees for the performance, and mainly focuses on the generation and classification tasks and still requires to store past samples. 

Memory replay approaches would either use a generator \cite{Lifelong_VAE,LifelongUnsupervisedVAE,GenerativeLifelong,Generative_replay,LifelongTeacherStudent,LifelongVAEGAN,Lifelonginterpretable,LifelongMixuteOfVAEs,LifelongTwin} or a memory buffer \cite{GradientLifelong,TinyLifelong,FunctionalRegularisation} as a replay mechanism generating data, which is statistically consistent to the previously learnt knowledge. Continual Unsupervised Representation Learning (CURL) \cite{LifelongUnsupervisedVAE} is a memory replay method which trains a latent generative model in order to replay data consistent with the previously learnt information. CURL expands its architecture for the inference component but not for the decoder. This can lead to catastrophic forgetting when learning certain tasks. 

Besides these three directions of research, there are other methods such as \cite{AdversarialContinualLearning,MixtureOfVAEs,DeepMixtureVAE} which create network architectures consisting of a shared module and other multiple task-specific modules. The shared module would not change its parameters too much, while the task-specific modules would only update their parameters when learning certain tasks. These approaches may guarantee the full performance on the previously learnt tasks but would still require knowing the number of tasks and the task labels during both training and testing phases. In this paper, we assume that our lifelong learning model does not know the exact number of tasks to be learnt while the task boundaries are only provided during the training phase. Although some methods can be used in a task-free manner \cite{LifelongUnsupervisedVAE}, they are still limited to learning a sequence of tasks from a single domain. 

\section{Methodology}

\subsection{Problem setting}

The lifelong artificial learning systems aim to learn a sequence of tasks, where each time we have a training set $D_S = \{ ({\bf x}_i,y_i)\}_{i=1}^N$ of $N$ paired instances of data ${\bf x}_i$, considered as images, and their corresponding labels $y_i$. Let us consider the training of a model (classifier or generator) ${\cal M}_t$, sequentially, with $t$ tasks, each defined by the training set and the testing set $D_T$. The learning goal of ${\cal M}_t$ is to make precise predictions for classification tasks on all testing data sets $\{D^1_T,\dots,D^t_T \}$ after the training with a sequence of sets $\{D^1_S,\dots,D^t_S \}$. When considering the unsupervised learning setting, the learning goal of ${\cal M}_t$ is to learn meaningful data representations without having any labels.

\subsection{The lifelong 
infinite mixture (LIMix) model}
\vspace{-1pt}

In this section, we first introduce a mixture of deep learning networks for unsupervised learning, and then extend this framework for a supervised setting. Let us define a deep learning mixture of $K$ components at the $t$-th task learning:
\begin{equation}
\begin{aligned}
p\left( {{\bf x},{\bf z} \mid \Theta , {\pi _1}, \ldots ,{\pi _K}} \right) = \sum\limits_{j = 1}^K {\pi_j}p_{\theta_j} \left( {\bf x}\mid{\bf z} \right)p\left( {\bf z} \right),
\label{mixObject}
\end{aligned}
\end{equation}
where $\Theta  = \left\{ \theta_1,\dots,\theta_K \right\}$ are the parameters of components. ${\bf x} \in \mathcal{X}$ and ${\bf z} \in \mathcal{Z}$ are the observed and latent variables where $\mathcal{X}$ and $\mathcal{Z}$ are the input and latent space. Each $p_{\theta_j}( {\bf x}\,|\,{\bf z} )$ is implemented as a Gaussian distribution ${\cal N}(g_{{\theta _j}}({\bf z}),\Sigma)$ with $\Sigma$ considered as a diagonal matrix, and $g_{\theta_j}$ represents the deterministic mapping that maps ${\bf z}$ into the mean of $p_{\theta_j}({\bf x}\,|\,{\bf z})$, implemented as a deep learning network \cite{MixtureOfVAEs,DeepMixtureVAE}. $\pi_j$ is the mixing parameter for the $j$-th component. $p(\bf z)$ is the prior, implemented by the normal distribution. One approach for training this model is to maximize the marginal likelihood of $p({\bf x},{\bf z}\mid \Theta,\Omega,{\pi _1},\dots,{\pi _K})$ as:
\begin{equation}
\begin{aligned}
& p\left( {\bf x}^1_1,\dots,{\bf x}^M_{N_M} \mid \Theta ,\pi_1, \ldots ,\pi_K \right) = \\ & \prod\limits_{t = 1}^M \prod\limits_{n = 1}^{N_t} \int \sum\limits_{j = 1}^K \pi_j p_{\theta_j}\left( {\bf x}_n^t \mid {\bf z} \right)  p\left( {\bf z} \right)\, \mathrm{d} {\bf z}\, \label{mix_likelihood},
\end{aligned}
\end{equation}
\noindent
where $M$ and $N_i$ are the total number of tasks and the number of data samples considered for each $i$-th task, $i=1,\ldots,M$. This optimization problem is intractable in practice since we cannot access data samples from previous tasks with associated training sets $\{ D^i_S \mid i=1,\dots,t-1 \}$ after learning the $t$-th task. Furthermore, by maximizing Eq.~\eqref{mix_likelihood} when learning only a single task would cause the mixture model to forget the information learnt previously, as the network parameters are updated to new values during the training with the data ${\bf x}_i^t \sim D^t_S$. To address this problem, we propose to adapt the number of components in the mixture according to the complexity of the tasks being learnt. The Dirichlet process is suitable for the selection and expansion mechanisms for the mixture \cite{InfiniteMixture}. In this paper we adapt a Dirichlet process by defining a probabilistic measure of similarity in order to be able to train the same mixing component with several tasks. 
We introduce an indicator variable $c_i^t$ for each ${\bf x}^t_i$ which indicates which component is assigned to ${\bf x}^t_i$. Estimating the mixing weights $\{ \pi_1, \ldots  \pi_K \}$ can be indirectly realized by the inference of the indicator variable, \cite{ProbabilityGMM}~:
\begin{equation}
\begin{aligned}
p\left(c_1^1,\ldots,c_{N_K}^K \mid \pi_1,\ldots, \pi_K \right) = \prod\limits_{j = 1}^K \pi_j^{N_j},
\end{aligned}
\end{equation}
 where $\{ \pi_1, \pi_2, \ldots, \pi_K \} \sim Dir\left({\bf a} \right), {\bf a}  =\{a/1,\cdot,a/K\}$ and $Dir({\bf a})$ is a symmetric Dirchlet distribution with ${\bf a}$ its parameter vector. Inferring a single indicator $c^t_i$ can be implemented by integrating over $\pi_j$ and allowing $K$ to increase to infinity, \cite{MixtureGaussianProcess}~:
\begin{equation}
\begin{aligned}
p \left(c_i^t =j \mid c^t_{-i},a \right) = \frac{n_{-i,j}}{n-1 + a},
\end{aligned}
\label{belonging}
\end{equation}
\noindent where $n_{-i,j}$ is the number of samples that are associated with the $j$-th component, excluding ${\bf x}_i^t$, where the subscript $-i$ denotes all indices except $i$. Eq.~\eqref{belonging} represents the probability for the $i$-th data sample to be associated with the $j$-th component of the mixture. The hyperparameter vector ${\bf a}$ can influence the prior probability of assigning a sample to a new component and the total number of components after training \cite{MixtureGaussianProcess}. However, this probability does not infer properly $c^t_i$ given that it does not evaluate the consistency of the new sample ${\bf x}^t_i$ with the information learnt by each component. Comparing the previously learned knowledge with the incoming data is useful for selecting the most suitable mixture component in order to be updated, or for adding a new component to the mixture. In this paper, instead of comparing the new sample with those already stored  \cite{DirichletLifelong,MixtureGaussianProcess}, we propose to incorporate the knowledge learned by each component for estimating $n_{-i,j}$ in order to consider the similarity between the prior knowledge and the new sample~:
\begin{equation}
\begin{aligned}
{n_{ - i,j}} = \left( {n - 1} \right) \times  {\frac{{{e^{\left( {1/{{\bf K}_{i,j}}} \right)}}}}{{ \sum\nolimits_{q = 1}^K {{e^{\left( {1/{{\bf K}_{i,q}}} \right)}}}  + e^{\left( 1/{\cal V} \right)}}}} \,,
\label{Select_Equ}
\end{aligned}
\end{equation}
where ${\mathcal V}$ is a constant controlling the expansion of the mixture model and
\begin{equation}
{\bf K}_{i,j} = \left|F\left({\bf x}_i^t \mid c_i^t,\theta_j,\omega_j \right) - F\left({\bf x}'_{i,j} \mid c_i^t,\theta_j,\omega_j\right) \right|
\label{Knowledge_eq1}
\end{equation}
and $F(\cdot \,|\, c_i^t,\theta_j,\omega_j)$ is the log-likelihood function. $n$ is the total number of samples, and ${\bf x}_{i,j}'$ is the $i$-th sample generated by the component $j$. We evaluate  ${\bf K}_{i,j}$ between the log-likelihood of the new sample ${\bf x}_i^t$ and the log-likelihood of the generated sample ${\bf x}_{i,j}'$, estimated by the $j$-th component. If ${\bf K}_{i,j}$ is very small, then ${\bf x}^t_i$ has a high likelihood to be assigned to the $j$-th component. The probability of generating a new component and of assigning the indicator variable $c^t_i$ to ${\bf x}_i^t$ is then defined as:
\begin{equation}
\begin{aligned}
p\left({c_i^t} =K+1\mid c^t_{ - i},a\right) =\frac{{a + \left(n-1\right)  \left(\frac{{\exp \left(1/{\mathcal V} \right)}}{Z}\right)}}{{n - 1 + a}}\,,
\end{aligned}
\end{equation}
where $Z=\sum\nolimits_{q = 1}^K e^{(1/{{\bf K}_{i,q}})}  + e^{(1/{\cal V})}$ is denominator of \eqref{Select_Equ}. 
\noindent \textBF{Determining the indicator for a new task.} By using the inferring indicator variable for all samples when learning the last given $t$-th task, is computationally intensive. We also know that data samples from a database, characterizing a certain task, share similar features. We consider only the calculation of a single indicator variable for a task, after each task switch, while the indicator variables for all data samples within a task are identical. Suppose that we have finished the $(t-1)$-th task learning, we would like to infer the indicator for the $t$-th task. Firstly, we randomly select a group of samples $\{{\bf x}^t_1,\dots,{\bf x}^t_{n_G}\}$ from the $t$-th training set and then calculate the probability of each ${\bf x}^{t}_i$ belonging to each $j$-th component $p(c^t_i = j \mid c^t_{ - i},a), \forall i = 1,\ldots,n_G$, where $n_G$ is the group size. Then we define the indicator $c^t$ for the $t$-th task by~:
\begin{equation}
\begin{aligned}
{c^{t}} = \mathop {\arg \max }\limits_{j = 1,\dots,K + 1} \frac{1}{{{n_G}}}\sum\nolimits_{i = 1}^{{n_G}} {p\left( {{c^t_i} = j \mid {c^t_{ - i}},a} \right)}\,,
\end{aligned}
\end{equation}
Once $c^t$ is determined, at the $t$-th task learning we only update the parameters of the chosen component instead of updating the whole model $\Theta$, by maximizing the sample log-likelihood $\log \int p_{\theta_{c^{t}}}({\bf x}\,|\,{\bf z})p({\bf z})\, \mathrm{d}{\bf z}$ which is intractable in optimization since we need integrating over ${\bf z}$. Similar to \cite{VAE}, we introduce to maximize a lower bound to the sample log-likelihood by using a variational distribution $q_{\omega_{c^{t}}}({\bf z}\,|\,{\bf x})$ at $t$-th task learning~:
\begin{equation}
\begin{aligned}
\log {p_{{\theta _{{c^{r}}}}}}({\bf{x}}) \ge \; & {\mathbb{E}_{{q_{{\omega _{{c^{t}}}}}}({\bf{z}}\mid{\bf{x}})}}\left[ {\log {p_{{\theta _{{c^{t}}}}}}({\bf{x}}\mid{\bf{z}})} \right] \\&- {D_{KL}}\left[ {{q_{{\omega _{{c^{t}}}}}}({\bf{z}}\mid{\bf{x}}) \mid\mid p({\bf{z}})} \right],
\end{aligned}
\end{equation}
where the right-hand side is our log-likelihood function 
$F(\cdot \,|\,{c^{t}},\theta_{c^{t)}},\omega_{c^{t}})$, called the Evidence Lower Bound (ELBO), used for training the model and the evaluation of Eq.~\eqref{Knowledge_eq1}.
$p_{\theta_{c^{t}}}({\bf x}\,|\,{\bf z})$ and $q_{\omega_{c^{t}}}({\bf z}\,|\,{\bf x})$ are decoding and encoding distributions, implemented by the network $g_{\theta_{c^{t}}}\colon \mathcal{Z} \to \mathcal{X}$ and $f_{\omega _{c^{t}}}\colon \mathcal{X} \to \mathcal{Z}$, respectively, where the subscript denotes the component index. In Lemma 4 from Appendix-F of the Supplementary Material (SM), we show how LIMix can infer across domains by the selection process.

\subsection{Learning prediction tasks}

In this section, we extend LIMix model for prediction tasks. Conditional VAEs \cite{LearnCGM} is one of the most used generative models for predictive tasks, defined by:
\begin{equation}
\begin{aligned}
\log p_{\varsigma_{c^t}}\left(y\mid{\bf x}\right) &\ge {\mathbb{E}_{{q_{{\omega _{c^t}}}}\left({\bf z}\mid{\bf x},y\right)}}\left[ {\log {p_{{\varsigma _{c^t}}}}(y\mid{\bf x}, {\bf z})} \right] -\\& {D_{KL}}\left({q_{{\omega _{c^t}}}}\left({\bf z}\mid{\bf x},y\right) \mid\mid {p_{{\varsigma_{c^t}}}}\left({\bf z}\mid{\bf x}\right)\right).
\label{item1}
\end{aligned}
\end{equation}
For classification, $y$ belongs to the discrete domain (one-hot vector), and $p_{\varsigma_{c^t}}(y\,|\,{\bf x},{\bf z})$ is implemented as a classifier. We represent $p_{\varsigma_{c^t}}({\bf z}\,|\,{\bf x})$ as ${\cal N}(0,{\bf I})$ in Eq.~\eqref{item1} for reducing the model size, and this results in the objective function:
\begin{equation}
\begin{aligned}
\mathcal{L}_{\rm{P}} 
 = \; & \mathbb{E}_{{q_{\omega_{c^t}}}({\bf z}\mid{\bf x},y)}\left[ \log {p_{{\varsigma_{c^t}}}}(y\mid{\bf x}, {\bf z}) \right] \\
& - D_{KL} (q_{\omega_{c^t}}({\bf z}\mid{\bf x},y) \mid\mid p({\bf z}))\,.
\label{classifier}
\end{aligned}
\end{equation}
We also require each component to learn a generator in order to overcome forgetting when reusing a selected component to model more than one task. Therefore, we define the ELBO for the generative model, $p_{\theta_{c^t}}({\bf x}\,|\, y)$ as~:
\begin{equation}
\begin{aligned}
\mathcal{L}_{\rm{G}}
= \; & \mathbb{E}_{{q_{\omega_{c^t}}}({\bf z}\mid{\bf x},y)} \left[ \log p_{\theta_{c^t}}({\bf x}\mid {\bf z},y) \right] \\
& - D_{KL}(q_{\omega_{c^t}}({\bf z} \mid {\bf x},y) \mid\mid p({\bf z}))\,.
\label{generator}
\end{aligned}
\end{equation}

Each component in the classification setting has three models $p_{{\varsigma _{c^t}}}(y \,|\,{\bf x}, {\bf z}),{q_{\omega_{c^t}}}({\bf z} \,|\,{\bf x},y),p_{\theta_{c^t}}({\bf x} \,|\, y, {\bf z})$. The likelihood function $F(\cdot \,|\, c^t,\theta_{c^t},\omega_{c^t})$ for the classification is only $\mathcal{L}_G$ and the main objective function for optimizing the $c^t$-th component is to maximize $\mathcal{L}_{\rm{G}} + \mathcal{L}_{\rm{P}}$. In practice, we optimize $\mathcal{L}_{\rm{P}}$ and $\mathcal{L}_{\rm{G}}$ separately, in the same mini-batch. In Appendix-J from SM, we provide the framework for applying LIMix for Image-to-Image Translation tasks and in Appendix-M we provide the experimental results. 

\subsection{Training a compressed Student model}

In order to reduce the complexity of LIMix, we propose to share most parameters of the generator and the inference models through a joint network, where the parameters $\theta_i = \{ \theta_S, \tilde{\theta}_i \}  $ and ${\omega _i} = \{ \omega_S, \tilde{\omega}_i\} $ of each component consists of the shared part $\{ \theta_S, \omega_S \} $ and the individual part $\{ \tilde{\theta}_i, \tilde{\omega}_i \}$ for each component. The mixing components are built on top of the shared component. We also train a compressed Student model under the unsupervised learning only, aiming to embed knowledge from LIMix into one latent space which supports interpolation across multiple domains. The Student shares the same network architecture with the component and is trained using the knowledge distillation (KD) loss along with the sample log-likelihood at the $t$-th task learning~:
\begin{equation}
\begin{aligned}
\mathcal{L}_{stu} = \underbrace{{\mathbb{E}_{S_{t,X}}}\log p_{\theta _{stu}}\left( {\bf{x}} \right)}_{\text{Log-likelihood}}
 + \underbrace{\sum\limits_{i = 1}^K {\mathbb{E}_{\mathbb{P}_{\theta_i}}}\log p_{\theta_{stu}}\left( {\bf{x}} \right) }_{\text{knowledge distilltion}}
\end{aligned}
\end{equation}
where ${\theta _{stu}} = \{ {\theta _S},{\tilde \theta _{stu}}\} $ and ${\tilde \theta _{stu}}$ is the individual set for the Student. $\log p_{\theta_{stu}}({\bf x})$ is estimated by ELBO and $\mathbb{P}_{\theta_i}$ is the distribution modelled by the $i$-th component in LIMix. More details together with the LIMix model diagram are provided in Appendix-J from SM. Additionally, this paper does not focus on the improvement of KD and we find that the Student is weaker than LIMix. This is theoretically explained in Appendix-I.2 from SM.

\section{Theoretical analysis for lifelong learning}

In this section, we first provide the theoretical analysis for the proposed infinite mixture model that does not grow its architecture during the lifelong learning. In this case, the model uses GRM to overcome the catastrophic forgetting and is seen as a single model represented by ${\cal M}(\theta ,\varsigma ,\varphi)$, consisting of a generator ${g_\theta }\colon \mathcal{Z} \to \mathcal{X}$ and a classifier $h_\varsigma \colon \mathcal{X} \to \mathcal{Y}$, where $\mathcal{Y}$ is an output space, which is $\{-1,1\}$ for binary classification and $\{1,2,\dots,n'\},n'>2$ for multi-class classification. We assume that the model also contains a task-inference network $U_\varphi \colon \mathcal{X} \to \mathcal{T}$ where $\mathcal{T}$ is the task domain. We then provide theoretical guarantees for the convergence of LIMix. Eventually, we analyse the forgetting behaviour of existing methods and the trade-off between the model's performance and complexity.

\subsection{Preliminaries}

\begin{definition} \emph{(Approximation distribution).}
Let us define a joint distribution $\tilde{S}^t$ approximated by the generator $g_{{\theta}^t}$ and the classifier $h_{\varsigma^t}$ of 
${\cal M}(\theta^t,\varsigma^t,\varphi^t )$ trained on a sequence of sets $\{ D_S^1,\ldots,D_S^t\}$. We assume that we have a perfect task-inference network $U_\varphi$, which can exactly predict the task label for a given sample ${\bf x}$. With the optimal task-inference network $U_\varphi$,\vspace{-1pt} we can form several joint distributions $\{ \tilde{S}_1^t,\ldots,\tilde{S}_t^1\} $ where each $\tilde{S}_i^{(t-i+1)}$ is made up of a set of samples where each paired sample is drawn by using the sampling process\vspace{-1pt} $\{{\bf{x}},h_{\varsigma^t}({\bf{x}})\} \sim {{\tilde S}^t}\;\text{if}\;{U_\varphi }({\bf{x}}) = i$. We use the superscript $(t-i+1)$ in $\tilde S_i^{(t-i+1)}$ to denote that $\tilde S_i^1$ is refined for $(t-i+1)$ times through the GRM processes after the $t$-th task learning.\vspace{-2.3pt} We further use $\tilde{S}_{i,\mathcal{X}}^{(t - i+1)}$ to denote the marginal distribution of $\tilde{S}_i^{(t - i + 1)}$. $\tilde S_{i}^n$ is formed by the samples from $D_S^i$ $\text{if } n = 1$, otherwise, by samples drawn from $G_{\theta^n}$ and $h_{\varsigma^n}$ with the optimal task-inference network.
\end{definition}
\begin{definition} \emph{(Data distribution across tasks).}
Let $S_i$ represent the joint distribution characterizing the probabilistic representation for the testing set in the $i$-th database $D_T^i$, and $S_{i,\mathcal{X}}$ is its marginal distribution along ${\mathcal{X}}$.
\end{definition}
\begin{assumption}
\label{assumption1}
We assume $\tau \colon \mathcal{Y} \times \mathcal{Y} \to [0,1]$ be a symmetric and bounded loss function $\forall (y, y') \in {\mathcal{Y}^2},\tau (y,y') \le {M'}$  and $\tau ( \cdot, \cdot)$ obeys the triangle inequality, where ${ M'}$ is a positive number. 
\end{assumption}
\vspace{-10pt}
\begin{definition} \emph{(Discrepancy distance).} For two given joint distributions $\tilde{S}_i^{(t - i + 1)}$ and $S_i$ over $\mathcal{X} \times \mathcal{Y}$ and $\tau :\mathcal{Y} \times \mathcal{Y} \to [0,1]$ is a loss function satisfying Assumption~\ref{assumption1}. Let $h, h' \in \mathcal{H}$ be two classifiers, where $\mathcal{H}$ is the space of all classifiers, and we define the discrepancy distance $\Psi$ between the two marginals $\tilde S_{i,\mathcal{X}}^{(t - i + 1)}$ and $S_{i,\mathcal{X}}$ as:
\begin{equation}
\begin{aligned}
\Psi \left(\tilde S_{i,X}^{(t - i + 1)},S_{i,X} \right)
 = & \mathop {\sup }\limits_{\left( h,h' \right) \in \mathcal{H}^2} \bigg| \mathop \mathbb{E}\limits_{\tilde S_{i,X}^{(t - i + 1)}} \left[ {\tau \left( {h'\left( {\bf{x}} \right),h\left( {\bf{x}} \right)} \right)} \right] \\&
-
{ \mathop \mathbb{E}\limits_{{S_{i,X}}} \left[ {\tau \left( {h'\left( {\bf{x}} \right),h({\bf{x}})} \right)} \right]} \bigg|.
\end{aligned}
\vspace{-5pt}
\label{discrepency}
\end{equation}
\end{definition}

\begin{definition} \emph{(Empirical risk).}
For a given loss function $\tau \colon \mathcal{Y} \times \mathcal{Y} \to [0,1]$ and a joint distribution $S_i$, we form an empirical set where we draw each paired sample as $\{{\bf{x}}_j^i,y_j^i\} \sim S_i$. the empirical risk ${\rm{R}}(h,{S_i})$ for a given classifier $h \in \mathcal{H}$, is evaluated by $n$ number of independent runs.
\begin{equation}
\begin{aligned}
{\rm{R}}\left(h,S_i\right)=\frac{1}{n}\sum\limits_{j = 1}^n \tau \left(h\left({\bf x}_j^i\right),y_j^i\right).
\end{aligned}
\end{equation}
\end{definition}

\subsection{Risk bounds for lifelong learning}

The discrepancy distance, defined through Eq.~\eqref{discrepency}, was used to derive generalization bounds for domain adaptation methods \cite{DomainRepresentation,domainRegression,domainTheory} and also for matching generated and real data distributions in the GAN's discriminator during training \cite{GAN_Maximum,MMDGAN,GMMN}. In the following we derive the risk bound for the lifelong learning model based on the discrepancy distance. The main idea for analyzing the degenerated performance of the model is to evaluate the risk between the target and the dynamically degenerated source distribution caused by the retraining process using GRMs. In this case, the errors accumulated when learning each task can be measured in an explicit way. 
\begin{theorem}
\label{theorem1}
    Let $S_i$ and $\tilde S_i^{(t - i + 1)}$  be two joint distributions over $\mathcal{X} \times \mathcal{Y}$. Let $h_i = \arg \min_{h \in \mathcal{H}} {\rm{R(}} h,S_i {\rm{)}}$ and $\tilde{h}_i^{(t - i + 1)} = \arg \min_{h \in \mathcal{H}} {\rm{R(}}h, \tilde{S}_i^{(t - i + 1)}{\rm{)}}$ represent\vspace{-1pt} the ideal classifiers for $S_i$ and $\tilde S_i^{(t - i + 1)}$, respectively, where $\mathcal{H}$ \vspace{1pt}is the classifier space. By satisfying Assumption~\ref{assumption1}, we have:
\begin{equation}
\begin{aligned}
{\rm{R}}\big(h,S_i\big) &\le {\rm{R'}}\big(h,\tilde h_i^{(t - i + 1)},\tilde{S}_i^{(t - i + 1)} \big) \\&+ \Psi \big(S_{i,X},\tilde S_{i,X}^{(t - i + 1)} \big) + \sigma \big(S_i,\tilde{S}_i^{(t - i + 1)}\big)
\end{aligned}
\label{theorem1_equ1}
\end{equation}
where the optimal combined error is represented by \\
\begin{equation}
\begin{aligned}
\sigma (S_i,\tilde{S}_i^{(t - i + 1)})&={\rm{R'}}(h^*_i,{h_i},{S_i})\\& + {\rm{ R'}}({h_i},\tilde h_i^{(t - i + 1)},\tilde S_i^{(t - i + 1)})
\end{aligned}
\end{equation}
and 
\begin{equation}
{\rm{R'}}(h_i^*,h_i,S_i) = {\mathbb{E}_{{\bf x} \sim {S_{i,X}}}}\tau (h_i^*({\bf{x}}),{h_i}({\bf x}))\,,
\end{equation}
where $h_i^*$ is the true labeling function for $S_i$.
\end{theorem}

We provide the proof in the Appendix-A from SM. This theorem provides a way to measure the gap on the risk bound for the model, after learning the $t$-th task, but does not provide any insight on how the previously learnt knowledge is forgotten. The following theorem provides an explicit way to measure the accumulated errors when learning a certain task.
\begin{theorem}
\label{theorem2_section}
Let $\tilde S_i^{(t - i + 1)}$ be the joint distribution over $\mathcal{X} \times \mathcal{Y}$, and $\tau$ be a loss function which satisfies Assumption~\ref{assumption1}. The accumulated errors in the knowledge associated with a previously learnt, $i$-th task, after learning a given $t$-th task, can be defined as~:
\begin{equation}
\begin{aligned} 
&{\rm{R}}(h,{S_i}) \le {\rm{R'}}(h,\tilde h_i^{(t - i + 1)},\tilde S_i^{(t - i + 1)}) +\\ & \sum\limits_{k = 0}^{t - i} {\left( {\Psi (\tilde S_{i,X}^{(k)},\tilde S_{i,X}^{(k + 1)})
+ \sigma (\tilde S_i^{(k)},\tilde S_i^{(k + 1)})} \right)}, 
\label{Equ_theorem2}
\end{aligned}
\end{equation}
where the last term of the right hand side (RHS) is expressed as~:
\begin{equation}
\begin{aligned} 
\sigma (\tilde S_i^{(k)},\tilde S_i^{(k + 1)}) &= {\rm{R'}}(\tilde h_i^{(k)},\tilde h_i^{*(k)},\tilde S_i^{(k)})\\& +{\rm{ R'}}(\tilde h_i^{(k)},\tilde h_i^{(k + 1)},\tilde S_i^{(k + 1)})\,,
\label{theorem2_equ0}
\end{aligned}
\end{equation}
\vspace{-10pt}
\end{theorem}
where we use $\tilde{S}_i^{({\rm{0}})}$ to represent $S_i$ for simplicity. The proof is provided in the Appendix-B from SM. Theorem~\ref{theorem2_section} provides the analysis of forgetfulness of the previously learnt knowledge in the model ${\cal M}(\theta^t ,\varsigma^t ,\varphi^t)$, while learning the $t$-th task. When $i$ is small, {\em i.e.} a task which was learnt during one of the initial training stages, then the accumulated terms $\sum\nolimits_{k = 0}^{t - i} {\tau (\tilde S_{i,X}^{(k)},\tilde S_{i,X}^{(k + 1)}) + \sigma (\tilde S_i^{(k)},\tilde S_i^{(k + 1)})}$ lead to larger errors. This explains that ${\cal M}(\theta^t ,\varsigma^t,\varphi^t)$ would tend to forget the tasks learnt earlier during its lifelong learning process. We visualize this forgetfulness process in Figure~\ref{theory_image}.
\begin{lemma}
\label{lemma1}
Let us consider that we have Assumption~\ref{assumption1}, then the accumulated error after learning the probabilistic representations of all databases after $t$-th task learning is:
\begin{equation}
\begin{aligned}
&\sum\limits_{i = 1}^t {{\rm{R}}(h,{S_i})} \le \sum\limits_{i = 1}^t \bigg( {{{\rm{R'}}(h,\tilde h_i^{(t - i + 1)},\tilde S_i^{(t - i + 1)}) }} +\\
& {\sum\limits_{k = 0}^{t - i } {\left( {\Psi (\tilde S_{i,X}^{(k)},\tilde S_{i,X}^{(k + 1)}) + \sigma (\tilde S_i^{(k)},\tilde S_i^{(k + 1)})} \right)} } \bigg)\,.
\label{theorem3_2}
\end{aligned}
\end{equation}
\end{lemma}
We consider Theorem~\ref{theorem2_section} and sum up the accumulated errors from Eq.~\eqref{Equ_theorem2} for learning $t$ tasks and this results in Eq.~\eqref{theorem3_2} (Appendix-C from SM). Lemma~\ref{lemma1} shows that minimizing the discrepancy distance $\Psi (\tilde S_{i,X}^{(k)},\tilde S_{i,X}^{(k + 1)})$ between the generated distribution approximated by the model and the target distributions, when learning each task, plays an important role in the improvement of the performance. However, the accumulated errors of the model will increase significantly when increasing the number of new tasks to be learnt. The following lemma shows how a mixture or ensemble model can address this problem and improve the performance.

\begin{lemma}
\label{lemma2}
Let us consider Assumption~\ref{assumption1} and assume that we are training the LIMix model with $K$ components onto the $t$-th task learning. If $K = t$, then the accumulated errors of the infinite mixture model for all tasks is defined as:
\vspace*{-0.4cm}
\begin{equation}
\begin{aligned}
\sum\limits_{i = 1}^t {{\rm{R}}\left( h,S_i \right)}  
\le \; &  \sum\limits_{i = 1}^t \left( {{\rm{R'}}(h,\tilde{h}_i^1,\tilde S_i^1)} \right. \\
& \left.{
+ 
\Psi (S_{i,X},\tilde{S}_{i,X}^1) + \sigma (S_{i},\tilde{S}_{i}^1)} \right).
\label{Lemma_2}
\end{aligned}
\end{equation}
\vspace*{-0.2cm}
\end{lemma}

We provide the proof in Appendix-D from SM. Lemma~\ref{lemma2} provides the framework for an optimal solution for LIMix in which the lifelong learning problem is transformed into a generalization problem under multiple target-source domains where there is no forgetting error during the training. $h$ is implemented by the mixture of $\{h_{\zeta_1},\dots,h_{\zeta_K} \},h_{\zeta_i} \in \mathcal{H},i=1,\dots,K$ in LIMix and therefore the performance on each target domain is relying on the generalization ability of the associated component. In practice, the number of components is smaller than the number of tasks being learnt. We investigate a specific case in Appendix-D from SM. The following lemma provides an analysis of the relationship between the model's performance and complexity.

\begin{table*} 
\centering 
\scriptsize
\setlength{\tabcolsep}{2.36mm}{
\begin{tabular}{l c c c c cc c c c cc c c c cc} 
\toprule 
& \multicolumn{5}{c}{\textbf{MSE}}&
\multicolumn{5}{c}{\textbf{SSMI}}& \multicolumn{5}{c}{\textbf{PSNR}} \\ 
\cmidrule(l){2-6}\cmidrule(l){7-11}\cmidrule(l){12-16} 
\textbf{Datasets} & LGM & CURL & BE & LIMix &Stud&  LGM & CURL & BE & LIMix &Stud & LGM & CURL & BE & LIMix &Stud
\\ 
\midrule 
MNIST &129.93 &211.21&19.24 &26.66&176.82 &0.45&0.46&0.92&0.88&0.42&14.52&13.27& 22.57& 21.09&13.72
		\\
		Fashion &89.28 &110.60&38.81 &30.19&178.04 &0.51&0.44&0.61&0.76&0.37&15.82&14.89& 14.46& 21.25&8.81
		\\
		SVHN &169.55 &102.06&39.57 &35.07&146.70 &0.24&0.26&0.66&0.65&0.47&8.11&10.86& 18.90& 14.92&13.58
		\\
		IFashion &432.90 &115.29&36.52 &30.14&158.18 &0.26&0.54&0.75&0.79&0.43&9.04&15.51& 19.32& 20.26&14.17
		\\
		RMNIST &130.28 &279.47&25.41 &22.80&157.55 &0.45&0.29&0.88&0.90&0.43&14.51&10.84& 21.31& 21.81&14.18
		\\
\midrule
	\textBF{Average} &190.38 &163.72&31.91 & \textBF{28.97}&163.45 &0.38&0.39&0.76&\textBF{0.79}&0.42&12.40&13.07& 19.31& \textBF{19.86}&12.89 \\
\bottomrule 
\end{tabular}}
\vspace{3pt}
\caption{The performance of various models after the MSFIR lifelong learning.}
	\label{Unsupervised1}
	\vspace{-15pt}
\end{table*}

\begin{lemma}
\label{lemma3}
Let $B = \{ b_1,\dots,b_j \}$ represent the labels for the tasks that the corresponding distributions $\{ \tilde S_{{b_1}}^{(1)},..,\tilde S_{{b_j}}^{(1)}\}$ are accessed only once after lifelong learning. Let $B' = \{ b'_1,\dots, b'_n \}$ indicate which task was used for re-training more than once.\vspace{2pt} We also define a set $ \hat{B}  = \{ \hat{b}_1,\dots,\hat{b}_n \}$ that records how many times each task was used for re-training,\vspace{2pt} where $\hat{b}_i > 1$\vspace{-1pt} represents that the $b'_i$-th task has been retrained for $(\hat{b}_i-1)$ times $\tilde S_{{b'_i}}^{(1)} \to \tilde S_{{b'_i}}^{({{\hat b}_i})}$ where $\tilde S_{{b'_i}}^{({{\hat b}_i})}$ represents the corresponding probabilistic representation.\vspace{2.3pt} For a given mixture model, we have~:
\begin{equation}
\begin{aligned}
&\sum\limits_{i = 1}^t {{\rm{R}}\left( {h,{S_i}} \right)}  \le \sum\limits_{i = 1}^{{\rm{card}}(B)} \left( {{\rm{R'}}(h,\tilde h_{{b_i}}^1,\tilde S_{{b_i}}^1) +
\Psi ({S_{{b_i},X}},\tilde S_{{b_i},X}^1) }
\right. \\
& \left.
{+
\sigma ({S_{{b_i}}},\tilde S_{{b_i}}^1)} \right)  + \sum\limits_{i = 1}^{{\rm{card}}(B')} \bigg( {{\rm{R'}}(h,\tilde h_{{b_i'}}^{(t - {{\hat b}_i} + 1)},\tilde S_{{b_i'}}^{(t - {{\hat b}_i} + 1)})} 
\\&
{+
\sum\limits_{k = 0}^{{{\hat b}_i}-1} {\left( {\Psi (\tilde S_{{{b_i'}},X}^k,\tilde S_{{{b_i'}},X}^{(k + 1)}) + \sigma (\tilde S_{{{b_i'}}}^k,\tilde S_{{{b_i'}}}^{(k + 1)})} \right)} } \bigg)  = {{\rm{R}}_{{\rm{mixture}}}}
\end{aligned}
\label{lemma3_equ}
\end{equation}
where $t>1$ represents the number of tasks being learnt. ${\rm card}(\cdot)$ denotes the cardinal number in the set which satisfies ${\rm card}(B) + {\rm card}(B') > K$ and $0 \le {\rm{card}}(B) \le K,0 \le {\rm{card}}(B') \le K,{\rm{card}}(B') = {\rm{card}}({\hat B})$, where $K$ is the number of components used for training. The risk for a single learnt model $M({\theta ^t},\omega^t,\psi^t)$ is defined as ${{\rm R}_{{\rm{single}}}}$, as in Eq.~\eqref{theorem3_2}, while the risk for the mixture model, ${\rm R}_{\rm{mixture}}$ is the RHS of Eq.~\eqref{lemma3_equ}. From all these expressions we have ${\rm R}_{{\rm{single}}} \ge {\rm R}_{{\rm{mixture}}}$.
\vspace{-5pt}
\end{lemma}

The proof is in the Appendix-E from SM. Lemma~\ref{lemma3} does not explicitly indicate which task is associated with the information recorded by a specific component of the LIMix model. Nevertheless, we provide an explicit way to analyse the risk bound for the mixture model in a more practical way, where we vary the number of components and the number of times the model was trained for each task, according to different learning settings such as by considering the learning order of tasks or their complexity. If $B' = \varnothing$, then Eq.~\eqref{lemma3_equ} is reduced to Eq.~\eqref{Lemma_2}, meaning a smaller gap in the risk bound, while requiring additional memory. On the other hand, if ${\rm{card}}(B) = 1 \Rightarrow B = \{ t\} $, the RHS of Eq.~\eqref{lemma3_equ} becomes equal to that from Eq.~\eqref{theorem3_2}, meaning a large gap on the risk bound. We define the ratio $v = (K - {\rm card}(B))/(K - {\rm card}(B'))$ as an index which explains the trade-off between the model complexity and its performance. When $v$ increases, the model improves its performance while also increasing its complexity. In contrast, when $v$ is small, the model's complexity is reduced while it accumulates more error terms when learning new tasks. In the Appendix-G from SM, we use the proposed theoretical framework to analyse the forgetting behaviour for various models such as the classical GRM model \cite{Generative_replay}, a mixture model enabled with expansion mechanisms \cite{LifelongUnsupervisedVAE}, an ensemble model from \cite{BatchEnsemble} and an episodic memory model \cite{GradientEpisodic,ContinualMemorable}. We also extend Lemmas~\ref{lemma2} and \ref{lemma3} for analyzing the risk bounds for a model when changing the order of tasks in the Appendix-H from SM.

\begin{figure*}[ht]
	\centering
		\hspace{-5pt}
	\subfigure[]{
		\includegraphics[scale=0.30]{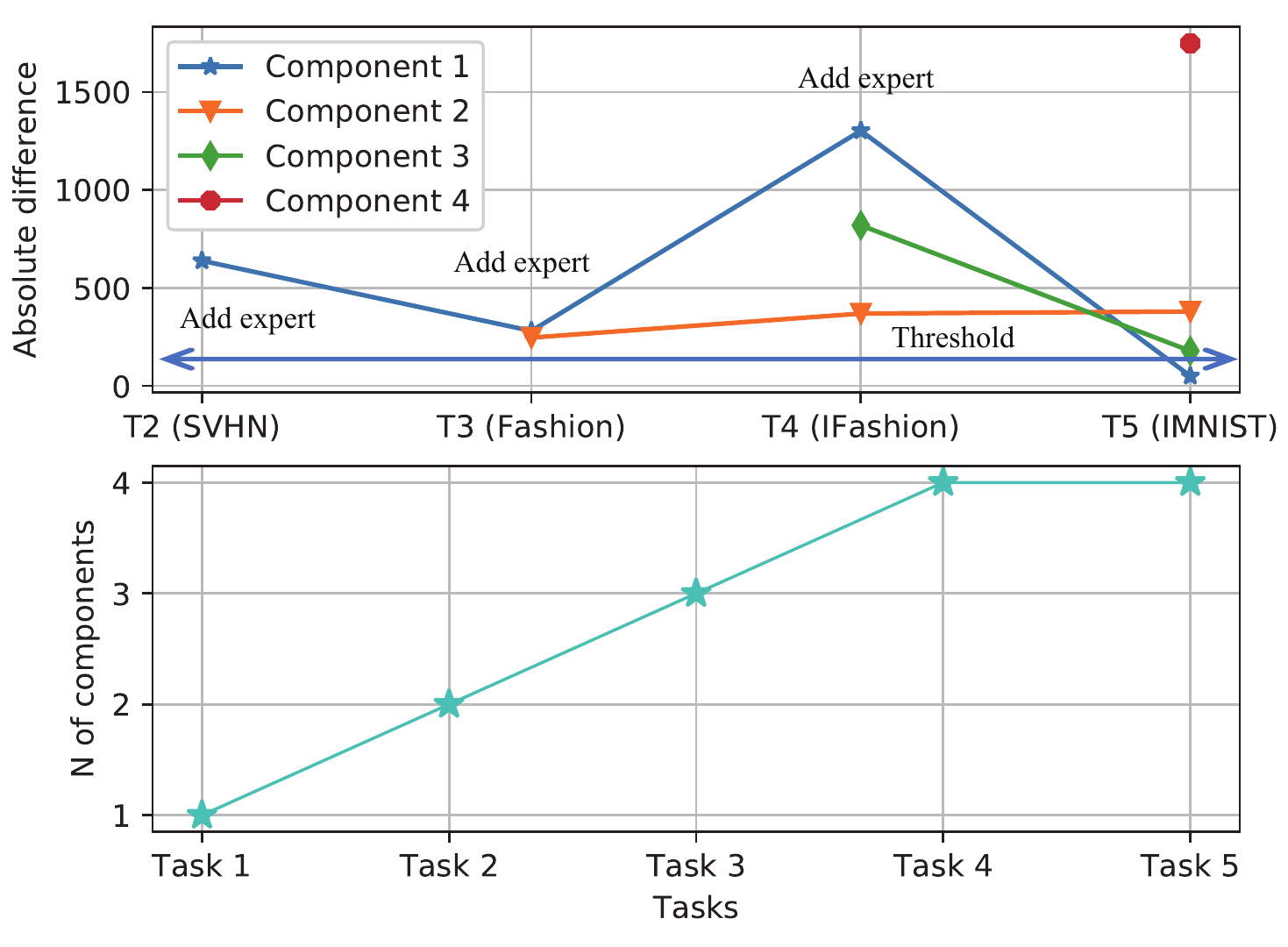}}
			\hspace{-10pt}
    \subfigure[]{
		\includegraphics[scale=0.30]{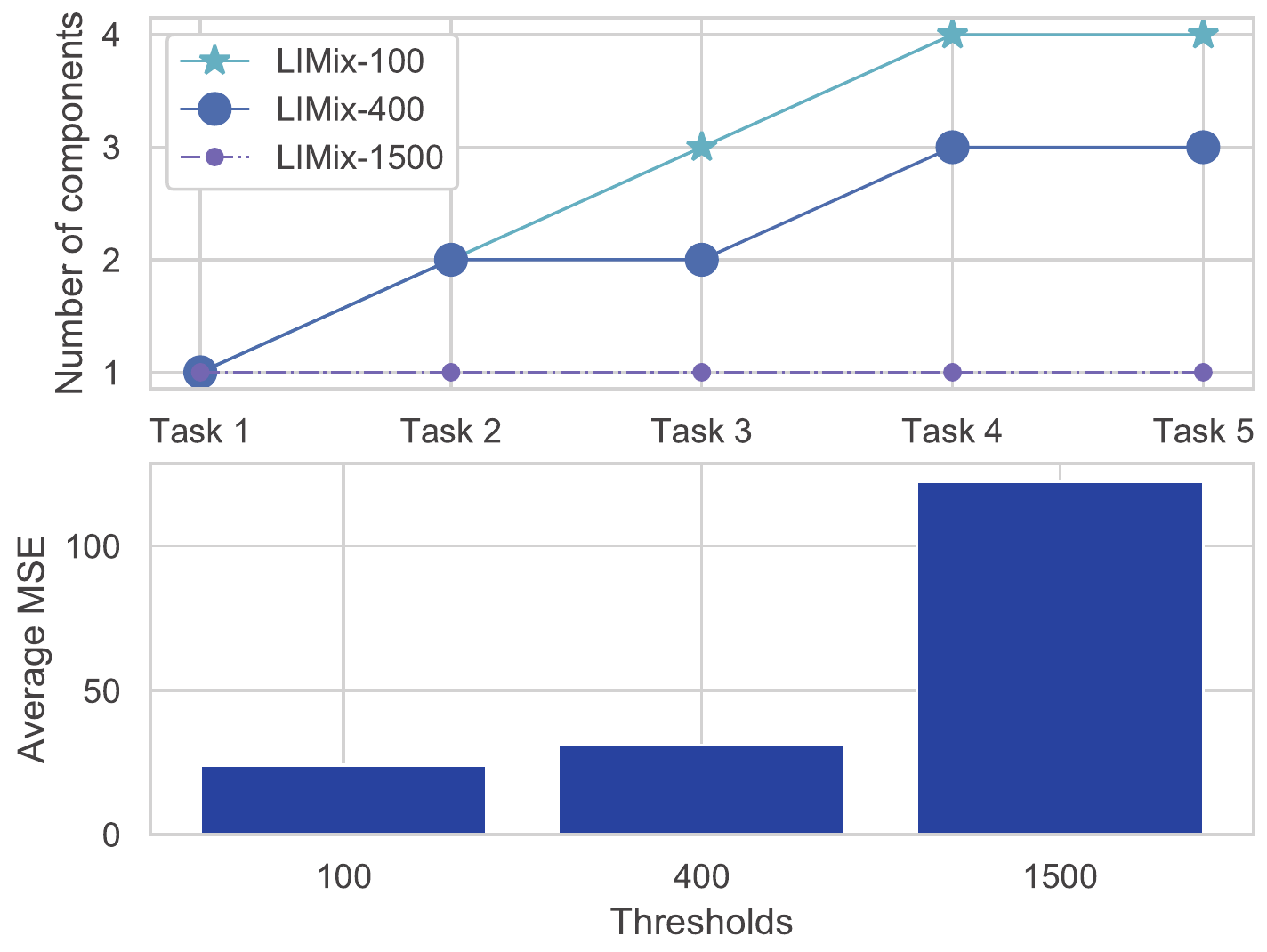}}
		\hspace{-7pt}
	\subfigure[]{
		\includegraphics[scale=0.30]{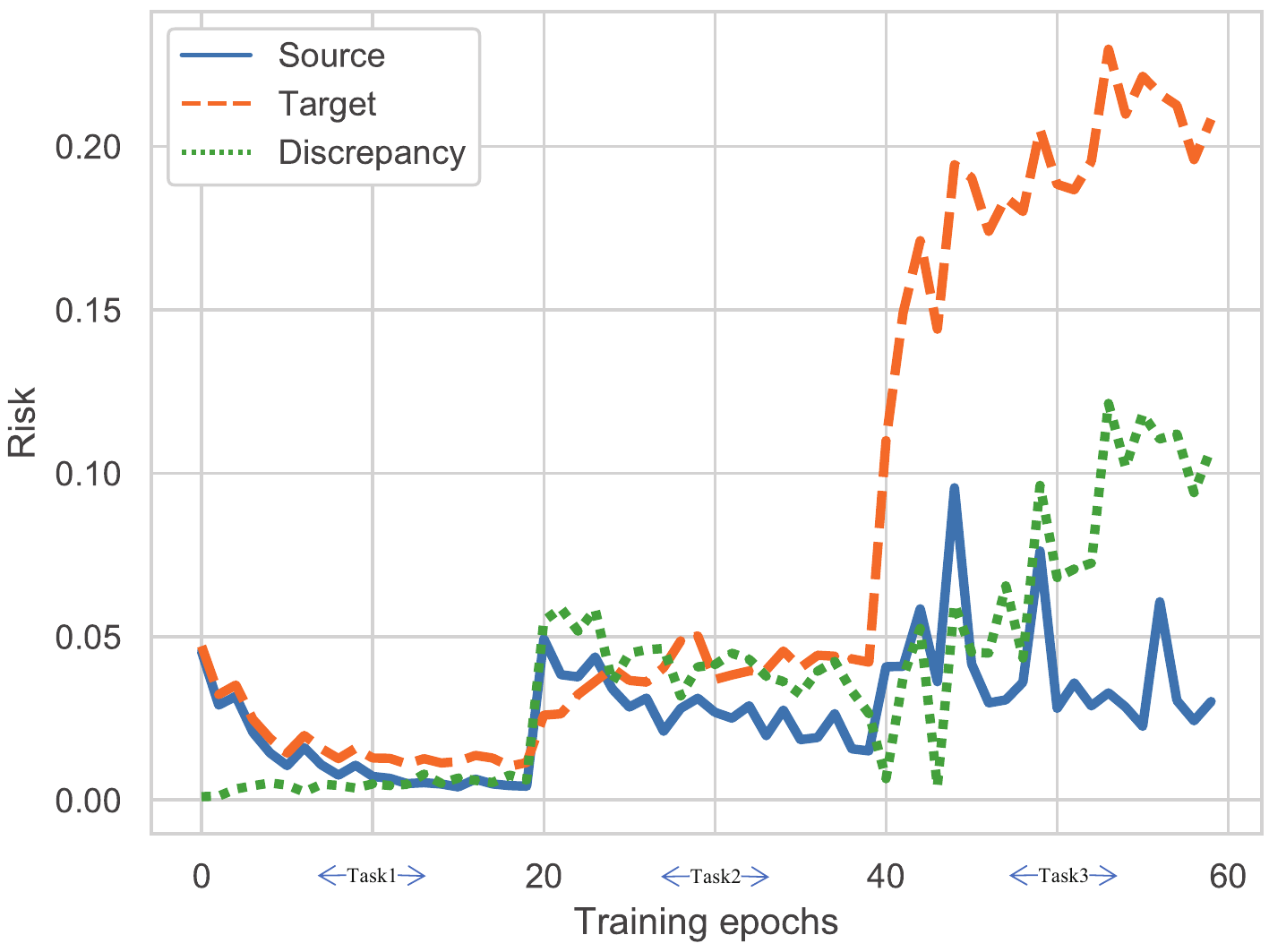}}
	\hspace{-7pt}
	\subfigure[]{
		\includegraphics[scale=0.30]{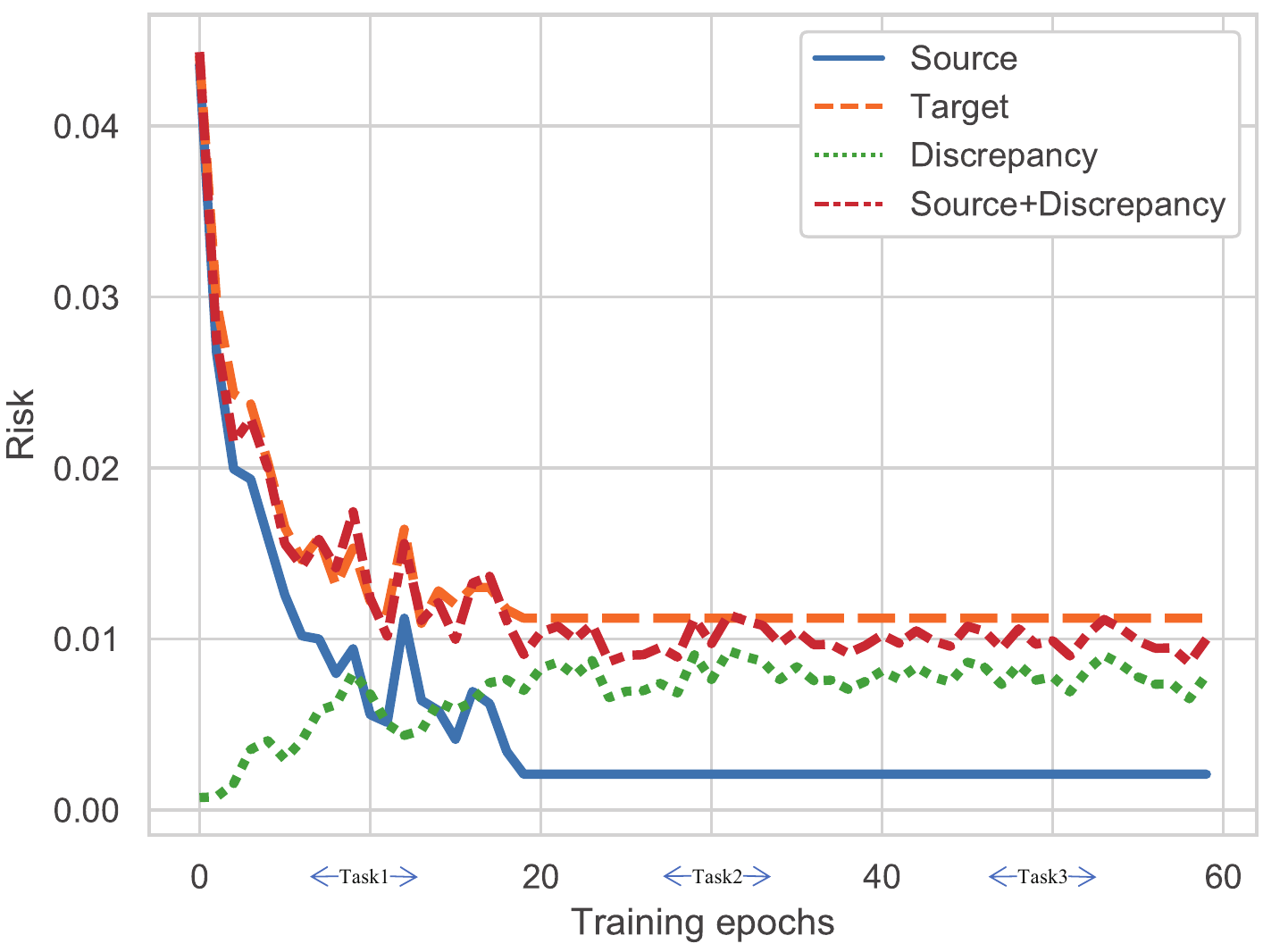}}
	\centering
	\vspace{-2pt}
	\caption{(a) Absolute difference on the log-likelihood and the number of components under the MSFIR lifelong learning. (b) The change of the model's performance and complexity when using different thresholds during MSFIR lifelong learning. (c) The risk and discrepancy for the first task (MNIST) by using LIMix without expansion. (d) The risk and discrepancy for MNIST by using LIMix with expansion.}
	\label{analysis}
	\vspace{-17pt}
\end{figure*}

\section{Experiments}
\subsection{Datasets and evaluation criteria}

We consider the following experimental settings~:
\begin{itemize}
\setlength{\itemsep}{2pt}
\setlength{\parsep}{0pt}
\setlength{\parskip}{0pt}

\vspace{-5pt}
\item For the unsupervised learning setting we create a sequence of learning tasks corresponding to the databases: MNIST \cite{MNIST}, SVHN \cite{SVHN}, Fashion \cite{FashionMNIST}, InverseFashion (IFashion) and Rated MNIST (RMNIST), and this learning setting is named MSFIR.
\item We add CIFAR10 \cite{CIFAR10} after MSFIR, as the last training task, resulting in MSFIRC sequence for supervised classification. All images are resized to $32\times 32 \times 3$.
\end{itemize}

\noindent \textbf{Evaluation criteria:}  In the classification tasks, we use the average accuracy over all tasks as the performance criterion. Although the proposed theoretical analysis is only used in prediction tasks, LIMix can also achieve good performance in the unsupervised setting where we use the Mean Squared Error (MSE), the structural similarity index measure (SSIM) \cite{Reconstruction_criteria} and the Peak-Signal-to-Noise Ratio (PSNR) \cite{Reconstruction_criteria} for the reconstruction quality evaluation.

\subsection{Unsupervised learning tasks}

We firstly evaluate various methods on MSFIR lifelong learning tasks and the results are provided in Table~\ref{Unsupervised1}. We compare our proposed LIMix model with three state of the art methods: LGM \cite{GenerativeLifelong}, CURL \cite{LifelongUnsupervisedVAE} and BatchEnsemble (BE) \cite{BatchEnsemble}. BE is designed to be used for classification tasks and we implement BE as an ensemble made up of VAE components, where each VAE has a tuple of trainable vectors built on the top layer of a neural network which does not update its parameters in the $k$-th task learning, $k=\{2,3,\dots,K\}$. We use large neural networks, containing more parameters, for LGM and BE, respectively, in order to ensure fair comparisons. The model size is provided in Appendix-L from SM. ``Stud'' denotes the performance of the Student model which is worse than LIMix because the Student learns its knowledge from the generation results of LMIX. Fig.~\ref{analysis}a shows the absolute difference on the log-likelihood between the incoming task and each component ${\bf K}_{i,j}$, as in Eq.~\eqref{Knowledge_eq1}, and the number of components derived during MSFIR lifelong learning. We can observe that the first component is reused when learning the fifth task (RMNIST) and LIMix expands to 4 components after LLL. We also evaluate the LIMix model when considering more complicated tasks in the Appendix-K.2 from SM.

\subsection{Classification tasks}

In this section we present the results when considering the lifelong learning of classification tasks. In order to use LGM \cite{GenerativeLifelong} for classification,
we continually train an auxiliary classifier on the real data samples from successive tasks and by also using paired data samples $\{{\bf x}_i,y_i\}_{i=1}^n$, where ${\bf x}_i$ is generated by either the Teacher or Student in LGM and each $y_i$ is inferred by the classifier during the last task learning. Table~\ref{classAcc} provides the LLL classification accuracy, where it can be observed that LIMix achieves the best results. Unlike in the image reconstruction results, CURL \cite{LifelongUnsupervisedVAE} also provides good results on the LLL of classification tasks. CURL uses a single decoder which is continually updated across multiple tasks and therefore leads to poor image reconstructions, which explains the difference between the reconstruction and the classification results for CURL.
\begin{table}
\vspace{4.5pt}
    \centering
\scriptsize
\setlength{\tabcolsep}{2.33mm}{
\begin{tabular}{@{}l c c c c c@{}} 
\toprule 
\textbf{Dataset}    & LGM \cite{GenerativeLifelong} & CURL \cite{LifelongUnsupervisedVAE} &BE\cite{BatchEnsemble}   & LIMix &MRGANs \cite{MemoryGAN} \\
\midrule 
MNIST&90.54 &91.30 &99.40 &91.16&91.24 \\
	SVHN&22.56 &62.05 &74.46 & 82.60&64.12 \\
	Fashion&68.29 & 79.18&88.95 &89.14&80.10  \\
	IFashion&73.70&82.51 &86.45  &88.70&82.19  \\
	RMNIST& 90.52 &98.56 &99.10 &98.80&98.30  \\
	CIFAR10& 57.43 &67.34 &52.48 & 54.66&67.19 \\
\midrule 
\midrule 
\textbf{Average}  & 67.17 &80.16 &83.47 &\textBF{84.18}&80.52 \\
\bottomrule 
\end{tabular}}
\vspace{3pt}
	\caption{Classification accuracy of various models after the MSFIRC lifelong learning.}
	\label{classAcc}
	\vspace{-15pt}
\end{table}

Although the proposed LIMix is mainly designed for cross-domain lifelong learning, we also apply LIMix to the continuous learning benchmarks, Permuted MNIST and Split
MNIST \cite{Continual_Learning} (see Appendix-K.4 from SM). Similar to \cite{ContinualMemorable}, we use a smaller network for the implementation of each component and perform five independent runs for calculating the mean and standard deviation. 
\begin{table}
\vspace{6pt}
    \centering
\scriptsize
\setlength{\tabcolsep}{3.15mm}{
\begin{tabular}{l c c c  } 
\toprule 
\textbf{Methods}   & Permuted MNIST & Split MNIST \\
\midrule 
DLP* \cite{LaplacePropagation} &$82\%$&$61.2\%$ \\
	EWC* \cite{EWC} &$84\%$&$63.1\%$ \\
	SI* \cite{Continual_Learning} &$86\%$&$98.9\%$ \\
	Improved VCL* \cite{ImprovedVCL}&$93.1\pm 1\% $& $98.4\pm 0.4\% $\\
	FRCL-RND* \cite{FunctionalRegularisation} &$94.2\pm 0.1\% $&$97.1\pm 0.7\% $ \\
	FRCL-TR* \cite{FunctionalRegularisation} &$94.3\pm 0.2\% $&$97.8\pm 0.7\% $ \\
	FROMP* \cite{ContinualMemorable} &$94.9\pm 0.1\% $&$99.0\pm 0.1\% $ \\ 
\midrule 
\midrule 
	LIMix  &\textBF{96.46}$ \pm 0.03\%$ (10 C) &\textBF{99.21}$ \pm 0.04\%$ (5 C) \\
		LIMix  &$88.78\%$ (7 C) &$96.77\%$ (4 C) \\ 
	LIMix  &$95.25\%$ (8 C)&$91.37 \%$ (3 C) \\ 
\bottomrule 
\end{tabular}}
\vspace{3pt}
	\caption{Results of continuous learning benchmark.}
	\label{classAcc2}
	\vspace{-15pt}
\end{table}
The results are provided in Table~\ref{classAcc2} where "*" means reporting the results from \cite{ContinualMemorable} and "5 C" means 5 components for LIMix. LIMix achieves the best performance for the optimal solution, validating Lemma~\ref{lemma2}, and would gradually loses performance when reducing the model size, as discussed in Lemma~\ref{lemma3}.

\subsection{Ablation study and theoretical results}

We evaluate the performance of the LIMix model when varying various hyperparameters and thresholds and the results are shown in Figures~\ref{analysis}a and \ref{analysis}b. The performance is improved by expanding the model's architecture, but this would increase its complexity, as discussed in Lemma~\ref{lemma3}. We provide additional results for the hyperparameter parameter setting in the Appendix-K.1 from SM. We investigate the theoretical results for Theorem~\ref{theorem2_section} and train a single model under MNIST, Fashion and SVHN (MFS) learning setting and evaluate the source-risk, target-risk and discrepancy on MNIST and present the results in Fig.~\ref{analysis}c. We can observe that the increase of the target-risk largely depends on the discrepancy instead of the source-risk which keeps stable when learning additional tasks. We also investigate the results for LIMix under MFS lifelong learning and the results are provided in Fig.~\ref{analysis}d, where "Source + Discrepancy" represents the source-risk plus the discrepancy on MNIST. The discrepancy does not increase for LIMix when learning additional tasks. Besides, "Source + Discrepancy" is very close to the target-risk and the gap to the target-risk is the combined error $\sigma ({S_1},\tilde S_1^1)$, according to Eq.~\eqref{Equ_theorem2}.

\begin{figure}[ht]
    \centering
    \hspace{-8.4pt}
	\subfigure[Target-risk]{
		\includegraphics[scale=0.29]{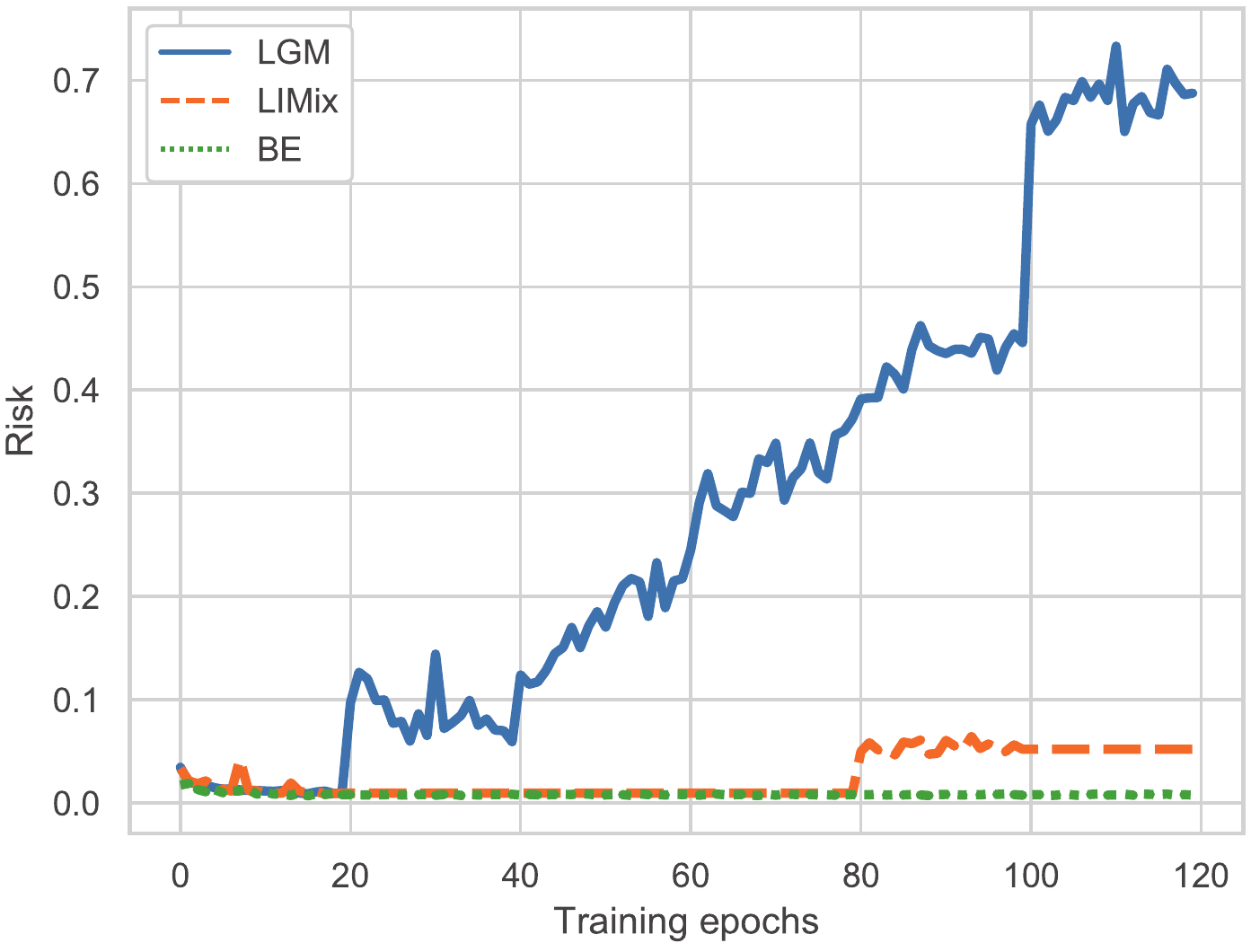}}
				\hspace{-4pt}
    \subfigure[Average target-risk]{
		\includegraphics[scale=0.29]{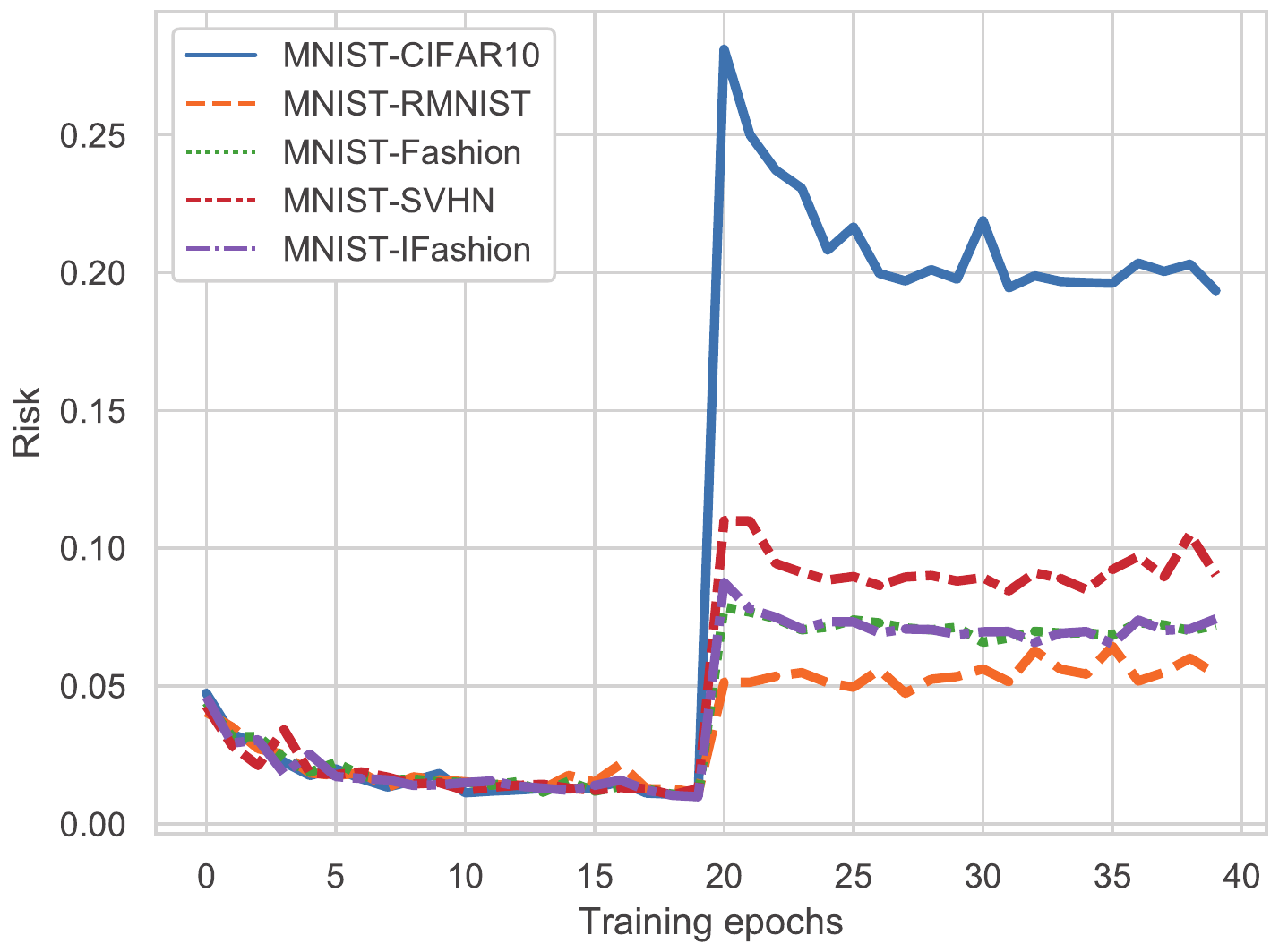}}
	\centering
	\caption{Evaluation of MNIST target-risk for other datasets.}
	\label{analysis2}
	\vspace{-10pt}
\end{figure}

We also investigate the target-risk for LIMix, BE, and LGM and the results are shown in Fig~\ref{analysis2}a, where the target-risk is evaluated on MNIST for each epoch. The target-risk for BE does not change across the LLL of six tasks, given that it does not accumulate any errors. LGM continually increases its target-risk, which is bounded by ${\rm R}_{\rm Single}$, adding an error term after learning each task. LIMix only increases the target-risk when it reuses a component, which was updated when learning RMNIST after MNIST. These results are explained by Lemma~\ref{lemma3} and further explored in Appendix-G from SM. We also estimate the average target-risk for all tasks (20 epochs for each task) when reusing a component trained on MNIST for learning a new task and the results are shown in Fig.~\ref{analysis2}b. We find that a single component would lead to a large degeneration in the performance when learning an entirely different task (CIFAR10) than a related task (RMNIST). This demonstrates that the proposed selection mechanism can choose an appropriate component minimizing the target-risk. Examples of image reconstruction, generation and Image to Image translation are shown in the Appendix-M from SM.

\vspace*{-0.1cm}
\section{Conclusion}

We propose a new theoretical analysis framework for lifelong learning based on the discrepancy distance between the probabilistic measures of the knowledge already learnt by the model and a target distribution. We provide an insight into how the model forgets some of the knowledge acquired during LLL, through the analysis of the model's risk. Inspired by the analysis, we propose LIMix model which performs better in cross-domain lifelong learning.

{\small
\bibliographystyle{ieee_fullname}
\bibliography{ICCV_final.bbl}
}

\end{document}